\pdfoutput=1

\documentclass[11pt]{article}

\usepackage{emnlp2021}

\usepackage{times}
\usepackage{latexsym}

\usepackage[T1]{fontenc}

\usepackage[utf8]{inputenc}

\usepackage{microtype}

\usepackage{url}
\usepackage[T1]{fontenc}
\usepackage{amsmath}
\usepackage{amsfonts}
\usepackage{graphicx}
\usepackage{multirow}
\usepackage{booktabs}
\usepackage{float}
\usepackage[caption = false]{subfig}
\usepackage[ruled,linesnumbered]{algorithm2e}
\usepackage{footnote}
\usepackage{colortbl}
\usepackage{xcolor}
\makesavenoteenv{table}
\makesavenoteenv{tabular}
\usepackage{makecell}

\title{End-to-End Conversational Search for Online Shopping\\ with Utterance Transfer}

\author{Liqiang Xiao$^{1}$\Thanks{ Work performed during internship at Amazon.},
Jun Ma$^{2}$, 
Xin Luna Dong$^{2}$, 
Pascual Martinez-Gomez$^{2}$, \\ 
{ \bf Nasser Zalmout$^{2}$, 
Chenwei Zhang$^{2}$, 
Tong Zhao$^{2}$, 
Hao He$^{1,3}$\Thanks{ Corresponding author},
Yaohui Jin$^{1,3 \dagger}$ 
}
\\
$^1$MoE Key Lab of Artificial Intelligence, AI Institute, Shanghai Jiao Tong University \\
$^2$Amazon.com, Inc.\\
$^3$State Key Lab of Advanced Optical Communication System and Network, \\ Shanghai Jiao Tong University\\
\small \texttt{ xiaoliqiang@sjtu.edu.cn}, 
\texttt{ \{junmaa, lunadong, gomepasc, nzalmout,} \\
\small \texttt{cwzhang, zhaoton\}@amazon.com}, \texttt{\{hehao, jinyh\}@sjtu.edu.cn}
}

\begin{document}
\maketitle
\begin{abstract}
    Successful conversational search systems can present natural, adaptive and interactive shopping experience for online shopping customers. 
    However, building such systems from scratch faces real word challenges from both imperfect product schema/knowledge and lack of training dialog data. 
    In this work we first propose \textbf{ConvSearch}, an end-to-end conversational search system that deeply combines the dialog system with search.  
    It leverages the text profile to retrieve products, which is more robust against imperfect product schema/knowledge compared with using product attributes alone. 
    We then address the lack of data challenges by proposing an utterance transfer approach that generates dialogue utterances by using existing dialog from other domains, and leveraging the search behavior data from e-commerce retailer. With utterance transfer, we introduce a new conversational search dataset for online shopping.
    Experiments show that our utterance transfer method can significantly improve the availability of training dialogue data without crowd-sourcing, and the conversational search system significantly outperformed the best tested baseline.
\end{abstract}

\section{Introduction}
    Search systems play significant roles in today's online shopping experience. In conventional e-commerce search systems, user interacts with the system through typing of keywords, followed by product clicks or keywords modifications, depending on whether returned product list matches with user expectation. The recent success of intelligent assistants such as Alexa, Google Now, and Siri enables user to interact with search systems using natural language. For online shopping in particular, it becomes alluring that users can navigate through products with conversations like traditional in-store shopping, guided by a knowledgeable yet thoughtful virtual shopping assistant.  
    
    However, building a successful conversational search system for online shopping faces at least two real world challenges. The first challenge is the \emph{imperfect product attribute schema and product knowledge}. While this challenge applies also to traditional search systems, it is more problematic for conversational search because the later depends on product attributes to link lengthy multi-turn utterances (in contrast to short queries in conventional search) with products. Most previous conversational shopping search work \cite{DBLP:conf/nips/LiKSMCP18,DBLP:conf/cikm/BiAZC19,DBLP:conf/aaai/YanDCZZL17} looks for the target product through direct attribute matching, assuming availability of complete product knowledge in structured form. In practice, this assumption rarely holds, and systems designed with this assumption will suffer from product recall losses.
    
    The second challenge is the \emph{lack of in-domain dialog dataset} for model training. Constructing a large-scale dialog dataset by crowd-sourcing from scratch is inefficient. Popular approaches include Machines-Talking-To-Machines (M2M) \cite{DBLP:conf/naacl/ShahHLT18}, which generates outlines of dialogs by self-play between two machines, and Wizard-of-Oz (WoZ) \cite{DBLP:journals/tois/Kelley84}, which collects data through virtual conversations between annotators. Note that both approaches require manually written utterances. In addition, a line of other work \cite{DBLP:conf/wsdm/Lei0MWHKC20,DBLP:conf/www/LuoSWLY20,DBLP:conf/cikm/BiAZC19} constructs conversations from the review datasets such as Amazon Product Data \cite{DBLP:conf/kdd/McAuleyPL15} and LastFM\footnote{\url{https://grouplens.org/datasets/hetrec-2011/}}, whereas usage of these datasets is limited to sub-tasks (e.g., dialog policy) due to the absence of user utterances. \citet{DBLP:conf/aaai/SahaKS18} collected a dialog dataset for fashion product shopping. However, the described method requires dozens of domain experts to manually create the dialog, and the dataset can hardly be generalized beyond fashion shopping given the lack of utterance annotations.

    To address the first challenge of imperfect attribute schema and product knowledge, we propose \textbf{ConvSearch}, an end-to-end conversational search system that deeply combines the dialog and search system to improve the search performance. 
    In particular, the Product Search module leverages both structured product attributes and unstructured product text (e.g. profile), where the product text may contain phrases matching with utterances when schema is incomplete or when a product attribute value is missing. Putting together, our system has the advantage of both reduced error accumulation along individual modules, and enhanced robustness against product schema/knowledge gaps.

    To address the second challenge of lacking in-domain dialog dataset, we propose a jump-start dialog generation method \textbf{M2M-UT} which 1) generates utterance from existing dialogues of similar domains (e.g., movie ticketing \cite{DBLP:conf/ijcnlp/LiCLGC17}), and 2) builds dialog outlines from e-commerce search behavior data, and fills them with the generated utterances. The proposed approach significantly reduces manual effort in data construction, and as a result we introduce a new conversational shopping search dataset \textbf{CSD-UT} with 942K utterances.
    Note that although the dialogue dataset construction focuses on shopping, the approach described here can be adapted for other task-oriented conversations as well, which we will leave it to future work. 
    Our contributions are summarized as follows: 
    \begin{itemize}
        \item We proposed an end-to-end conversational search system which deeply combines dialog with search, and leverages both structured product attributes and unstructured text in product search to compensate for incomplete product schema/knowledge.\footnote{The prototpye development, evaluation, and data set presented in this paper are independent from any existing commercialized chatbot system.}
        
        \item We proposed a new dialog dataset construction approach, which can transfer utterances from dialogs of similar domains and build dialogues from user behavior records. Using this new approach which significantly reduced manual work compared with existing approaches, we introduced a new conversational search dataset for online shopping.
      
        \item Extensive experiments show that our system outperforms evaluated competitors for success rate (SR@5) score.
    \end{itemize}

\section{Related Work}
    \paragraph{Conversational Search System}
    Conversational search task aims to understand user's search intents through multi-round conversational interactions, and return user the desired search item. Due to lack of annotated dialog utterances in particular for conversational search tasks, previous work either adopted rule-based utterance parsing or focused only on dialog policy. 
    \citet{DBLP:conf/aaai/YanDCZZL17} proposed a rule-based approach to cold-start online shopping dialog systems utilizing user search logs and intent phrases collected from community sites. 
    In another line of work, \citet{DBLP:conf/www/LuoSWLY20} and \citet{DBLP:conf/cikm/ZhangCA0C18} utilized Amazon review dataset, \citet{DBLP:conf/wsdm/Lei0MWHKC20} and \citet{DBLP:conf/nips/LiKSMCP18} revised user reviews from Yelp\footnote{\url{https://www.yelp.com/dataset/}} and LastFM\footnote{\url{https://grouplens.org/datasets/hetrec-2011/}}, all of which focused on the conversation policy without utterance understanding.
    As a comparison, in this paper we focused an end-to-end conversational search system, which fuses both utterance understanding and product search together through multi-task learning.
      
    \begin{figure*}[!t]
        \centering
        \includegraphics[width=\textwidth]{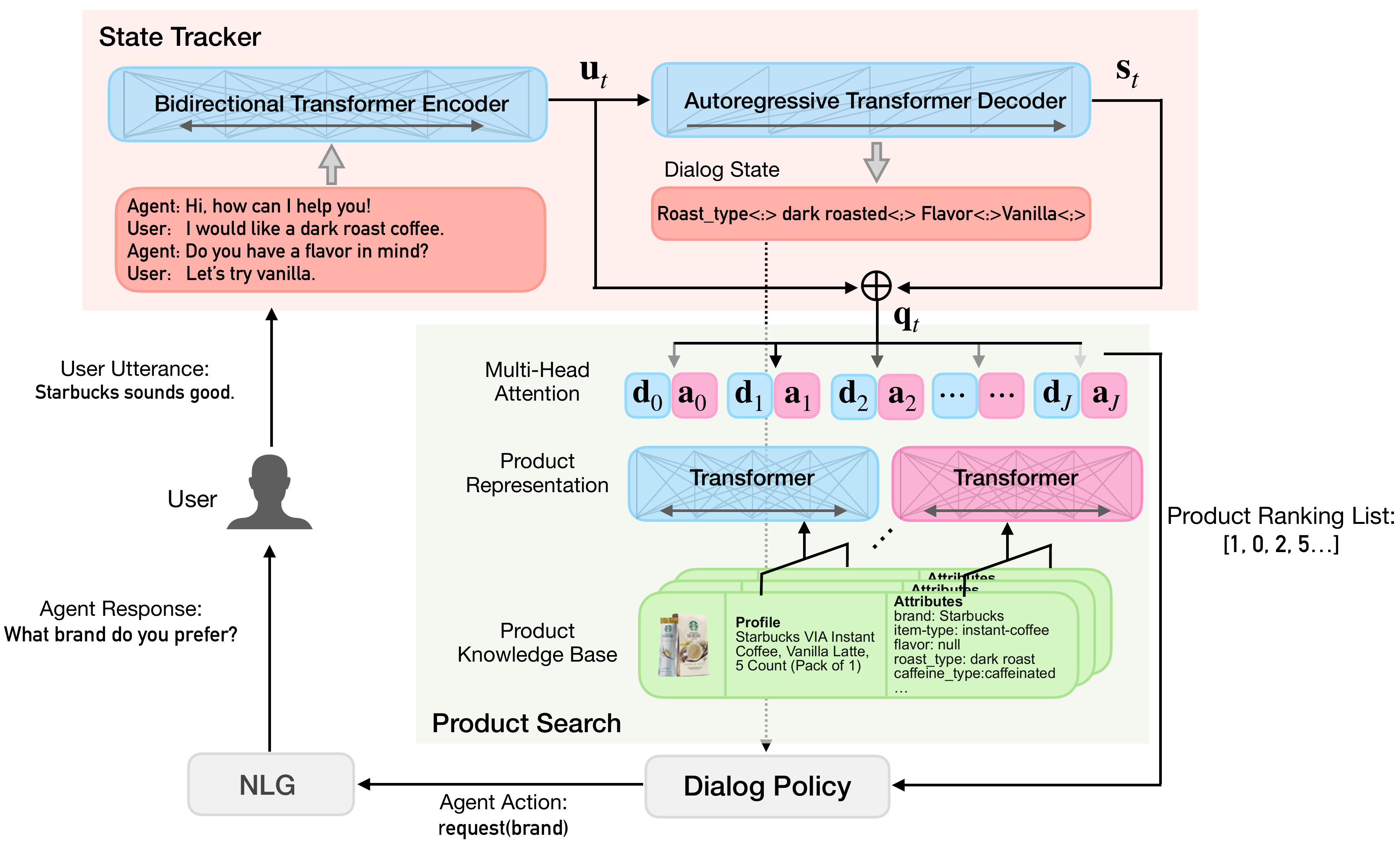}
        \caption{Illustration of our end-to-end conversational search system. State Tracker module takes utterances to predict dialog state $S_t$ using the seq-to-seq transformer. Product Search module matches products represented by transformers with query representations $\mathbf{q_t}$ using a multi-head attention mechanism. Dialog Policy module takes inputs from $S_t$, intent and ranked product list, and decides the responses. NLG module composes system responses as instructed by Dialog Policy, and displays them to user.}
        \label{neural_search_img}
    \end{figure*} 
    
    \paragraph{Constructing Dialog Dataset for Online Shopping}
    \citet{DBLP:conf/aaai/RastogiZSGK20} proposed a crowd-sourcing version of Wizard-of-Oz (WOZ) paradigm for collecting domain-specific corpora. In this system, users and wizards were given a predefined task to complete (e.g. find a Chinese restaurant in the North). To avoid the distracting latency, users and wizards were asked to contribute just a single turn for each dialogue. 
    \citet{DBLP:conf/aaai/SahaKS18} built a multi-mode dialog system for fashion with experts and in-house labors. They crawled 1 million fashion items from the web, hand-crafted taxonomy for items, identified the set of fashion attributes, and employed experts to write dialogs. The described methods were highly labor-consuming, and the published dataset did not contain attribute annotation on utterances, making it hard for utterance understanding model training. 
    The approach adopted by \citet{DBLP:conf/aaai/YanDCZZL17} mines phrases of shopping from community sites and uses crow-sourcing to label utterance intents. Although costing less labors, this work did not construct the full dialogs.
    As a comparison, in this paper we constructed the full shopping search dialogs through real user behavior data, with user utterances filled by transferring from existing dialogs of similar domains.
  
\section{End-to-End Conversational Search with Multi-task Learning}\label{sec-e2e}

    Our conversational search system, depicted in Figure \ref{neural_search_img}, consists of four major modules: State Tracker (ST), Product Search, Dialog Policy, and Natural Language Generation (NLG). The State Tracker module interprets the dialog content, outputs user intent, along with product attributes that the user is interested in. The Product Search module returns a list of products which are matching with tracked attributes user interested in. Based on output from State Tracker, the Dialog Policy manages agent response according to user intents and the candidate search result, and the NLG module transforms the response into natural language displayed to user.
    
    To address the challenge of imperfect product attribute schema/knowledge, our Product Search module leverages both structured product attributes and unstructured text.
    To mutually benefit from each other's learning, we integrate State Tracker and Product Search together through multi-task learning, and build an end-to-end trainable search system.
    
\subsection{State Tracker}\label{sec-nlu} 
    Unlike previous work that treats state tracking as a multi-label classification task \cite{DBLP:conf/acl/ZhuZFLTLPGZH20,DBLP:conf/icml/WenMBY17}, we redefine the state tracking task as a sequence-to-sequence problem. 
    As shown in Figure \ref{neural_search_img}, we link the slots and values of dialog state with special delimiter tokens, turning it into a sequence. 
    Then we employ a transformer network to translate dialog turns into state, which encodes the dialog lines with a bidirectional transformer encoder and generate state sequence autoregressively. 
    
    At each turn in the dialog, the State Tracker module outputs 1) the dialog \emph{state} $S$ and 2) the user utterance \emph{intent} $I$, where $S$ are attribute-values grouped by product attributes, representing the system's tracking of user's preferred search criteria, and \emph{intent} $I\in \mathbb{I}$ is an enumerable value from $\mathbb{I}=$\{\textit{request}, \textit{inform}, \textit{ask\_attribute\_in\_n}, \textit{buy\_n}\}.

    Formally, given a dialog at turn $t$, we have all history records of $\{R_0, U_0, R_1, U_1, \cdots, R_t, U_t, \}$, where $R_t$ and $U_t$ are \emph{system response} and \emph{user utterance} at turn $t$ respectively. We then use a transformer model \cite{DBLP:conf/naacl/DevlinCLT19} to predict the string $S_t$, the \emph{state} at $t$:
    \begin{equation}\label{eq-transformer}
        S_{t} = trans(concat(R_0;U_0;\cdots;R_{t-1};U_{t-1}))
    \end{equation}
    where $trans(\cdot)$ is the transformer, and $concat(\cdot)$ is a string concatenation function. For \emph{state} prediction, we use the loss function:
    \begin{equation}
        \mathcal{L}_s = - \sum_t \sum_{i} \log P(y^{i*}_t)
    \end{equation}
    where $y^{i*}_t$ denotes the ground-truth value for $i$-th item of output sequence at $t$-th turn.

    We also use an MLP layer to predict the \emph{intent} at $t$: 
    \begin{align}
        &\mathbf{z}_{t} = \mathbf{W}_I \cdot \text{ReLU}(\mathbf{u}_t) + b_I \\
        & P_t(I_i)       =  \text{softmax}(\mathbf{z}_{t})  
    \end{align}
    where $\mathbf{u}_t$ is the mean pooling of last layer output from the encoder in Equation (\ref{eq-transformer}), $W_I$ and $b_I$ are trainable parameters, and $P_t(I_i) $ represents the likelihood of \textit{intent} $I_i \in \mathbb{I}$  from user utterance at $t$-th turn. We use the following loss function for intent prediction: 
    \begin{equation}
        \mathcal{L}_I = - \sum_t \log P_t(I_i^*).
    \end{equation}
    where $I_i^*$ is the ground truth of \emph{intent} at turn $t$.

\subsection{Product Search}
   
    At each turn $t$, given current \emph{state}, Product Search module estimates the matching likelihood $P_t(p_j)$ for each product $p_j$, and then rank the products to be displayed to user (Figure \ref{neural_search_img}).

    \paragraph{Query Representation}
    We represent the product query as $\mathbf{q}_t=\mathbf{u}_t \oplus \mathbf{s}_t $, where $\mathbf{s}_t$ is the state representation obtained by mean pooling the last layer of decoder in Equation (\ref{eq-transformer}), and $\oplus$ denotes the vector concatenation operator. 
    
    \paragraph{Product Embedding} We represent $j$-th product  with $\mathbf{p}_j = \mathbf{d}_j \oplus \mathbf{a}_j$, where $\mathbf{d}_j$ is the mean pooling of the last encoding layer of: $trans(\text{description text of } p_{j})$, and $\mathbf{a}_j$ is the product attribute embedding. In particular, we obtain $\mathbf{a}_j$ by mean pooling the last layer of: $trans(\text{attribute sequence of } p_{j})$, where the attribute sequence is constructed as state sequence.

    The introduction of profile embedding $\mathbf{d}_j$ compensates for the missing matching clue when product schema is incomplete or attribute values are missing, since they may be extracted from product text.

    \paragraph{Search with Multi-Head Attention} We use multi-head attention mechanism to match query and products. At dialog turn $t$, we first calculate a product context vector $\mathbf{head}^k_t$ based on the glimpse operation \cite{DBLP:journals/corr/VinyalsBK15}: 
    \begin{align}
          & \mathbf{a}_{t}^{j}= \mathbf{v}_s^{k \top} \text{tanh}( \mathbf{W}^k_{p}\mathbf{p}_j + \mathbf{W}^k_{q}\mathbf{q}_t )  \label{attention_one}  \\
          & \mathbf{\alpha}_{t}= \text{softmax}(\mathbf{a}_{t}) \\
          & \mathbf{head}^k_t = \sum_{t} \mathbf{\alpha}_{j}^{t} \mathbf{W}^k_{p}\mathbf{p}_j
    \end{align}
    where $\mathbf{\alpha}_t$ are attention weights, and $\mathbf{W}^k_p, \mathbf{W}^k_q, \mathbf{v}^k_s$ are trainable parameters for head $k$. We then concatenate $K$ attention heads each with individual parameter sets, $\mathbf{head}_t = \oplus_{0 \leq k\leq K} \mathbf{head}^k_t$.
    
    We then form likelihood of product $p_j$ at $t$-th turn as:
    \begin{align}
            & \mathbf{e}^j_t = \mathbf{v}_p^{\top} \text{tanh}(\mathbf{W}_{p}' \mathbf{p}_j + \mathbf{W}_h \mathbf{head}_t) \label{attention_two} \\
            & P_t(p_j) = \text{softmax}(\mathbf{e}_t)
    \end{align}
    where $\mathbf{v}_p$, $\mathbf{W}_{p}'$, and $\mathbf{W}_h$ are trainable parameters.
    We use the following loss function for product search job:
    \begin{equation}
    \mathcal{L}_p = -  \sum_t \log P_t(p_j^*)
    \end{equation}
    where $p_j^*$ is the ground-truth of product. 
    Finally, we rank products with their likelihood, and return top products to Dialog Policy module for displaying. 
    
\subsection{Multi-task Learning}
    Our end-to-end training links all three tasks (\textit{state} prediction, \textit{intent} prediction and Product Search) together through multi-task learning:
    \begin{equation}
      \mathcal{L} = \alpha \mathcal{L}_s + \beta \mathcal{L}_I + \gamma \mathcal{L}_p
    \end{equation}
    where $\alpha, \beta, \gamma$ are tunable hyper-parameters. With multi-task learning, these three tasks can enhance each other with shared weights and back-propagated errors.
    
    The training data requires intent and attribute annotation for each utterance, and purchased products with product attributes and text profiles (optional) associated with each dialog.

\subsection{Dialog Policy and Natural Language Generation}
    During the conversation, the agent needs to propose additional attributes for user to narrow down the search. When triggered, we filter our product knowledge base using current \emph{state} $S$ to retrieve products matching with the criteria, then use EMDM (Entropy Minimization Dialog Management) \cite{DBLP:journals/taslp/WuLL15} to select the proposed attribute with maximum entropy among filtered products, and show user recommended narrowing down question. 
    
    The Natural Language Generation module translates the action decision from the Dialog Policy module to natural language, e.g. \emph{request(brand)} $\rightarrow$ \emph{Do you have a brand in mind?}. In this paper we simply use manually written agent templates.

\section{Dialogue Dataset Construction}\label{sec_transfer}
    We address the challenge of lack of conversational shopping search training data by proposing \textbf{M2M-UT}, a method that automatically constructs dialog datasets. Unlike previous works \cite{DBLP:conf/aaai/SahaKS18} that rely on crow-source to generate utterances, M2M-UT can automatically generates utterances with transfer. 
    
    We hypothesize that the conversation between the user and the shopping agent is guided by customer's \textit{intents} that 1) span user's utterance in natural language, and 2) change according to agent's responses. Therefore, our dataset construction has two steps: 1) we use utterance transfer (UT) to generate utterances from existing dialog datasets of similar domains, and 2) we generate the outline of dialog using customer browsing records using Machine Talking To Machine (M2M) \citet{DBLP:conf/aaai/SahaKS18}.
    
  \begin{figure}[!t]
        \centering
        \vspace{0.2in}
        \includegraphics[width=\linewidth]{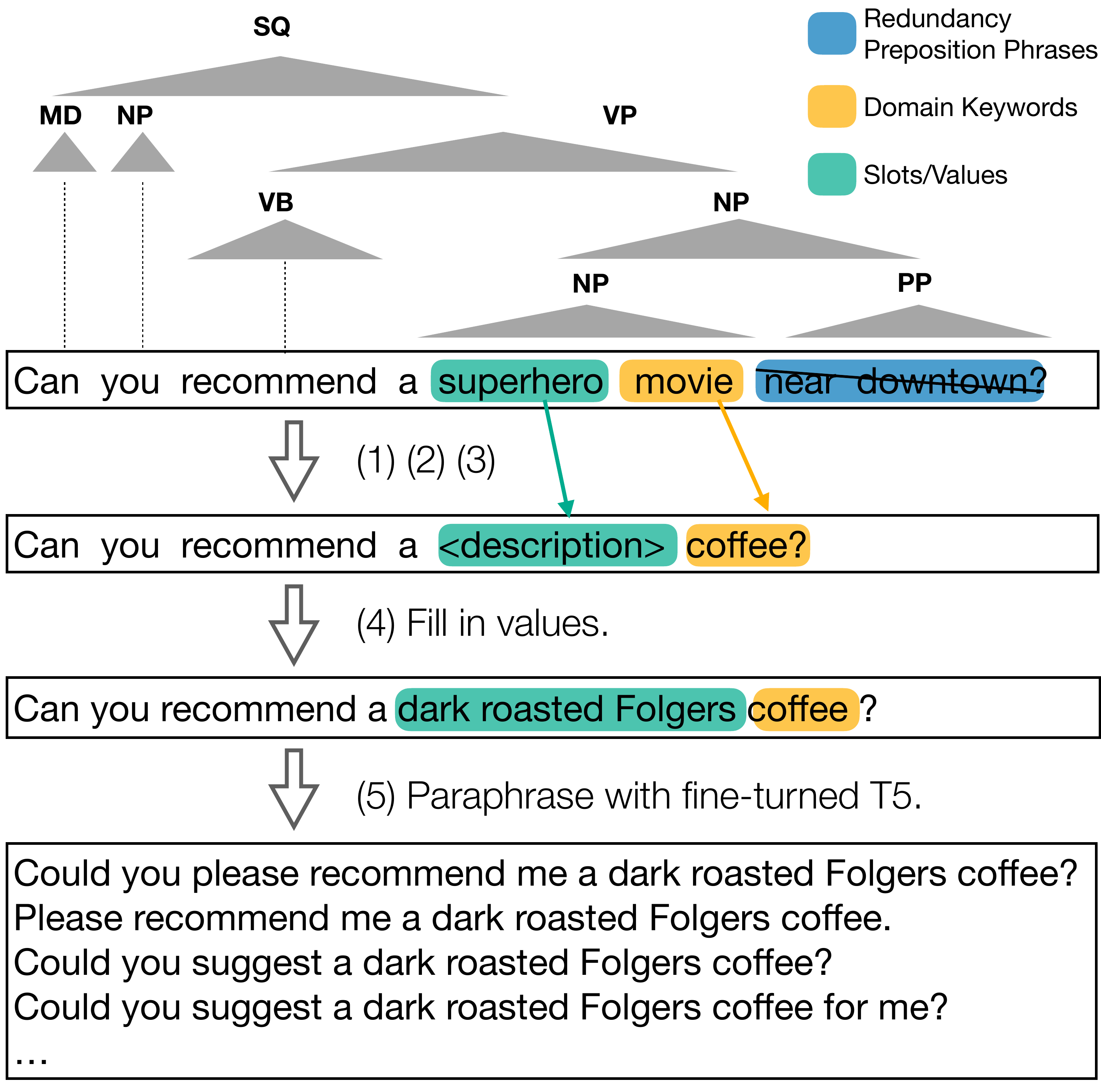}
        \caption{Utterance generation algorithm to generate variant utterances for coffee shopping domain. The utterance example employed in this figure is from MDC dataset \cite{li2018mschallenge}. An utterance is first transferred to our domain with the help of constituency parser and then paraphrased to enhance the variance.}
        \label{template_transfer}
    \end{figure}

    \subsection{Utterance Generation by Transfer}\label{ut}
        For utterance generation, widely used methods such as WoZ and M2M still require workers to create the various utterances, and thus are not easy to scale up in the shopping conversation application. We found that dialogues from existing task-oriented domains such as movie ticketing or restaurants reservation contain rich form of utterances similar to shopping, for example, ``... sounds good'' is seen from both movie ticketing and shopping conversations. We propose utterance transfer (\textbf{UT}), a novel approach that generates shopping utterances from related task-oriented domains.

        As shown in Figure \ref{template_transfer}, UT consists of five stages. (1) \textbf{remove redundant phrases}: we remove the redundant phrases that not commonly seen in online shopping (e.g. location and time) with syntax rules. We employ a constituency parser \cite{Kitaev-2018-SelfAttentive} to get the syntax tree of the sentence and remove the PPs (preposition phrases) and NPs (noun phrases) referring to location and time. (2) \textbf{replace values with slots}: we identify and replace values with slots according to the original dataset annotations. For example, in Figure \ref{template_transfer}, we identify value ``superhero'' using the annotation, and replace it as slot ``<description>''. This step turns a complete utterance into a template. (3) \textbf{keyword replacement}: we replace verbs and nouns with those from online shopping domain with rules, e.g. \emph{``movie''} to \emph{``coffee''} and \emph{``watch''} to \emph{``drink''}. (4) \textbf{fill slots}: we fill the slots with values according to user's action. (5) \textbf{paraphrase}: to augment the diversity of utterance, we use a fine-tuned T5 model \cite{DBLP:journals/jmlr/RaffelSRLNMZLL20} to paraphrase the utterance. 
 
    \paragraph{Paraphrase}
        One pitfall of utterances generated by templates and rules are the lack of  diversity, whereas real conversations usually contain various ways of expressing the same intents. As paraphrase can improve the performance of dialog system \cite{DBLP:conf/acl/GaoZOY20}, we employ a pre-trained neural paraphrase model to augment the variance of templates. Specifically, we use a T5 model (Text-to-Text Transfer Transformer)\footnote{https://github.com/ramsrigouthamg/Paraphrase-any-question-with-T5-Text-To-Text-Transfer-Transformer-} \cite{DBLP:journals/jmlr/RaffelSRLNMZLL20} that is fine-tuned on paraphrase dataset, Quora Question Pairs \footnote{https://www.quora.com/q/quoradata/First-Quora-Dataset-Release-Question-Pairs}.

  \begin{table}[!t]
        \centering
        \vspace{0.1in}
        \resizebox{\linewidth}{!}{
        \begin{tabular}{|l|l|}
        \hline 
        \multicolumn{2}{|l|}{\bf{Search Behavior Data}} \\
        \hline 
        \multicolumn{2}{|l|}{Search keywords: vanilla instant coffee packets} \\
         \multicolumn{2}{|l|}{
                \makecell[l]{Goal: \{flavor:vanilla, item\_type:instant-coffee,\\ brand:Folgers, roast\_type=medium roast,...\} } } \\
        \multicolumn{2}{|l|}{Purchase: ID=B074FLFKNV} \\
        \hline 
        \hline 
          \bf{Outline}   & \bf{Utterances} \\
        \hline
           S: greeting()              & Hello, what can I do for you?   \\ 
        \hline 
           U: \makecell[l]{inform(flavor=vanilla, \\ item\_type=instant-coffee)}  & \makecell[l]{Please find me vanilla instant\\ coffee packets.} \\ 
        \hline 
           S: request(brand)          & Do you have a brand in mind?    \\
       \hline 
           U: inform(brand=Folgers)   & Let's try Folgers. \\
       \hline 
           S: push(top\_5)            & \makecell[l]{I found you following products:\\  <Products List> }          \\ 
      \hline 
           U:\makecell[l]{ ask\_attr\_in\_n(roast\_type, \\index=2)} & \makecell[l]{What roast type is it in the \\ second image.} \\
     \hline 
           \makecell[l]{S: inform(roast\_type=\\medium roast)}  & It is medium roast.        \\
     \hline 
           U: buy\_n(index=2)                      & I will buy the second one. \\
    \hline 
           S: notify\_success()                 & Your order has been placed.  \\
        \hline 
        \end{tabular}
        }
        \caption{M2M-UT dialog generation. M2M-UT first generates the dialog outline with search behavior data, then translates it to utterances with the method illustrated in Figure \ref{template_transfer}. S and U denote System and User respectively. We use search keywords to generate the first user utterance.}
        \label{dialog generation}
    \end{table}

    \subsection{Dialog Generation}\label{sec_dialog_gen}
        Our online shopping dialog in conversational search is supported by the dialog \textit{outlines}, which consists of \textit{intent} and its parameters. For user utterance intents as shown in Table \ref{dialog generation} , their parameters are typically a list of product attributes with their values. For agent intents, parameters are either attribute values, or operation parameters that agent should execute with (e.g., push(top\_5)). Similar to the dialog system presented in Section \ref{sec-e2e}, we use \textit{state} to track agent's understanding of user's search criteria.
        
        We use real e-commerce search behavior data to supervise the construction of intent flow in the dialog. Each anonymous search session contains a query and the final purchased product. We first extract product attribute values from the search keywords as the initial attribute customer interested in, i.e., initial \textit{state}. We then follow M2M (Machine Talking to Machine) \cite{DBLP:conf/naacl/ShahHLT18} to generate the transition of dialogue outlines turn by turn. M2M runs in a self-playing manner by simulating the dialog with a user simulator and a system agent. We build an agenda-based user simulator initialized by the search behavior data, and use a finite state machine \cite{DBLP:books/daglib/0016921} as the system agent. 
        
        By comparing initial \textit{state} and the finally purchased product, we find that users were not always aware of the full search criteria at the beginning, therefore the dialog is constructed to simulate how agent helps user to fill the gap through attribute refinement. Specifically, as shown in Table \ref{dialog generation}, user starts with initial \textit{state} (e.g., flavor=vanilla). Given current \textit{state}, agent in the next turn proposes a new attribute (e.g., brand) using the policy EMDM (Entropy Minimization Dialog Management) \cite{DBLP:journals/taslp/WuLL15} to narrow down the search. User's response in the next turn will be based on attribute value of the purchased product (e.g., brand=Folgers), which also updates the \textit{state}. Then agent displays a list of products in the next turn (e.g., push(top\_5)). If purchased product appear in push list, user asks questions, commits purchase, and ends the dialog (successful). Otherwise agent proposes a new attribute, and continues the conversation. Dialog ends when length exceeds 20 (unsuccessful). 
        
        We finally translate the generated outline into natural language by using corresponding utterance templates generated after step (3) in Section \ref{ut}, and finalize the utterance following step (4) and (5) in Section \ref{ut}. After these steps, we have a complete shopping search dialog.
        
        
        

\section{Experiments}


\subsection{Datasets} 

Our dataset includes three parts: user search behavior data, dialogs, and product knowledge base. 

The user search behavior data is a collection of user search keywords and their final purchased products sampled from e-commerce platform. We applied the dialog generation method described in Section \ref{sec_transfer} on the coffee shopping domain. We leveraged the utterances from dataset MDC \cite{li2018mschallenge} and MMD \cite{DBLP:conf/aaai/SahaKS18} and transferred 4 intents from their domains (i.e. movie ticketing, restaurant reservation, fashion shopping), which generated 49,999 dialogs, with each of the dialog contains on average 18.85 turns (Table \ref{statistic_dialog}). In addition, we built up a gold-standard test set of 196 dialogs manually written by workers to evaluate the performance.

For the product knowledge base, we sampled 154,161 coffee products from the e-commerce platform. As shown in Table \ref{statistic_kb}, each product has a text profile with average 17.34 tokens and also the attribute-value pairs for 13 different attributes. The vacant ratio of values is 32.16\%, which indicates potential missing attribute values for products.

 \begin{table}[!tp]
        \centering
        \resizebox{0.8\linewidth}{!}{
        \begin{tabular}{|l|r|}
        \hline
        Metric                    & \textbf{CSD-UT}  \\
        \hline 
        \#Dialogs                 &   49,999      \\
        \#Total utterances        &   942,766     \\ 
        \#User utterances         &   471,383       \\
        Avg. \#Turns per dialog   &   18.85   \\
        Avg. \#Tokens per utterance   & 6.57     \\
        \hline
        \end{tabular}
        }
        
        \caption{Statistics of generated dialog dataset.}
        
        \label{statistic_dialog}
    \end{table}

    \begin{table}[!tp]
        \centering
        \resizebox{0.8\linewidth}{!}{
        \begin{tabular}{|l|r|}
        \hline
        \bf{Metric}                    & \bf{Product KB.}  \\
        \hline 
        \#Products                     & 154,161         \\ 
        \#Attributes                   & 13              \\
        Avg. \#Values per attribute    &  1111.13        \\
        Avg. \#Tokens per profile  &  17.34          \\
        Vacant ratio of values        &  32.16\%        \\
        \hline
        \end{tabular}
        }
        
        \caption{Statistics of product knowledge base.}
        
        \label{statistic_kb}
         
    \end{table}

\subsection{Settings}

    \paragraph{Hyper-parameters} 
     All the transformers used in experiment have 4 sublayers with hidden size of 256, and a word2vec \cite{DBLP:conf/nips/MikolovSCCD13} of 256 dimension is trained to initialize the embedding matrix. Our model used a vocabulary of 50257 entries for text embedding, and 14700 entries for attribute embedding. The models in experiments were trained with AdamW \cite{DBLP:journals/corr/abs-1711-05101} optimizer with the initial learning rate of 1e-4 and batch size of 16. The initial learning rate is selected based on validation loss. We used learning rate scheduler to cut the learning rate by half every time the performance drops and stop training once the performance has three straight drops. Our model was trained on a Nvidia Tesla P100 machine with 16G memory, and the strongest model (ConvSearch w/ Neural Search (attr.\&text.)) took 35 hours for convergence. For multi-task learning, we briefly set $\alpha, \beta, \gamma$ as 1. To save memory, we let the encoder of state tracker and encoder of profile share the parameters, and employed tf$ \cdot$idf to narrow down the search space into 400 products for product search module. 
    
    \paragraph{Evaluation Metrics}
    We use the success rate ($SR@t$) to measure the ratio of successful conversations, i.e. recommended the ground-truth item in $t$ turns. We set the max turn $t$ of a session to 5 or 10 and standardized the recommended list length as $5$. 
    Besides, we used recall, precision and F1 to evaluate the performance of state prediction, and reported the performance on slot and value respectively.  

\begin{table}[!tp]
    \centering
    \small
    \begin{tabular}{|l|cc|}
        \hline 
        \bf{Model}   &  \bf{SR@5} & \bf{SR@10}  \\
        \hline  
        TC-bot \cite{DBLP:conf/ijcnlp/LiCLGC17}       &    35.71     &   51.02  \\
        ConvLab-2 \cite{DBLP:conf/acl/ZhuZFLTLPGZH20} &    44.89     &   54.08 \\
        \hline 
        ConvSearch  &          &     \\
        \quad w/ Rule Search                    &   39.79     & 50.51   \\
        \quad w/ Neural Search (attr.)          &   46.42     & 57.14   \\
        \quad w/ Neural Search (text.)          &   48.47     & 59.69   \\
        \quad w/ Neural Search (attr. \& text.) &  \bf{51.53} & \bf{61.22} \\
        \hline 
    \end{tabular}
    \caption{Evaluation of the end-to-end system. attr. and text. denote attribute and product text respectively. The best score per metric is in bold. Our model outperforms the competitors by 6.64\%. Rule search employs direct attribute matching as traditional work.}
    
    \label{dialog_system}
\end{table}

\begin{table*}[!t]
    \centering
    \small
    \begin{tabular}{|l|ccc|ccc|ccc|}
        \hline 
        \bf{Model} & \multicolumn{3}{c|}{\bf{State-Attr}} & \multicolumn{3}{c|}{\bf{State-Value}} & \multicolumn{3}{c|}{\bf{Intent}} \\
        & R  & P   & F1  & R  & P   & F1   & R  & P   & F1\\         
        \hline 
e2e-Trainable \cite{DBLP:conf/eacl/Rojas-BarahonaG17} 
& 92.67 & 82.74 &  87.46  & 90.98  & 86.57 & 88.66 & 95.75 & 95.91 & 95.82    \\    

ZS-DST \cite{DBLP:conf/aaai/RastogiZSGK20} 
& 96.97 & 89.55 &  93.11  & 91.41 & \bf{87.70} & 89.51 & 96.43 & 97.89 & 97.15   \\   

LSTM + classification  & 92.34 & 89.72 & 91.01 & 88.97 & 82.31 & 85.51 & 95.65 & 94.26 & 94.94 \\ 
\hline 
State Tracker w/o Search  & 97.53 & 93.29 & 95.36   & 92.01 & 87.27 & 89.58 & \bf{99.73} &  99.68  & \bf{99.70}  \\  
State Tracker w/ Search   & \bf{98.15} & \bf{93.41} & \bf{95.72} & \bf{93.15} & 87.44 & \bf{90.20} & 99.70 & \bf{99.69}  & 99.69  \\  
        \hline 
    \end{tabular}
     
    \caption{Evaluation of state tracking task. R and P denote recall and precision.}
    
    \label{SLU_table}
\end{table*}

    \paragraph{Baselines}
    For state tracking task,  we compared against the following  baselines: e2e-Trainable \cite{DBLP:conf/eacl/Rojas-BarahonaG17} which encodes utterances with a convolutional neural network (CNN), ZS-DST \cite{DBLP:conf/aaai/RastogiZSGK20}, a Bert-based model which first judges the presence of each slot then the start and end location. We also constructed a baseline by replacing transformers in our system with one-layer LSTMs.  
    For the end-to-end system, we compared against two baselines: TC-bot \cite{DBLP:conf/ijcnlp/LiCLGC17}, a modulized neural dialogue systems for task-completion, and ConvLab-2\footnote{we employ the strongest setting they reported, BERTNLU+RuleDST+RulePolicy	+TemplateNLG.}\cite{DBLP:conf/acl/ZhuZFLTLPGZH20}, an open-source toolkit for building, evaluating, and diagnosing a task-oriented dialogue system. 

\subsection{End-to-End System Evaluation}
 Table \ref{dialog_system} shows the end-to-end task (success rate) comparisons, where our method outperforms baselines significantly by 6.64\%. This indicates the effectiveness of our end-to-end framework that deeply combines the dialog and search system, while ablation studies (last three rows in Table \ref{dialog_system}) also justify that leveraging both product text and attribute performs better than using only one of them.

\subsection{State Tracker Evaluation}
 Table \ref{SLU_table} shows the performance comparisons of state tracking task. It shows that our method outperforms all baselines in both \emph{state} prediction and \emph{intent} prediction tasks, which is because our state tracking task can better embed the context by concatenating the language of turns together. We also found State Tracker alone without Product Search task showed lower performance, suggesting the effectiveness of multi-task learning.

\begin{table}[!tp]
    \centering
    \small
    \begin{tabular}{|l|ccc|}
    \hline
    \bf{Model}           & \bf{R} & \bf{P} & \bf{F1} \\
    \hline 
    tf$\cdot$idf                       & 5.58       & 1.16  & 1.86         \\
    \hline
    Product Search w/attr.             & 15.53      & 3.10  & 5.16            \\
    Product Search w/text.             & 19.27      & 4.84  & 7.74           \\
    Product Search w/text. \& attr.    & \bf{26.20} & \bf{5.47}  & \bf{9.05}  \\
    \hline 
    \end{tabular}
     
    \caption{Independent evaluation of search task. This experiment shows the benefit of combining product text profile and attribute for search. attr. is abbreviation for product attribute. The best score per metric is in bold.}
     
    \label{neural_search_evaluation}
\end{table}

\subsection{Product Search Evaluation}\label{search_task_result}

Table \ref{neural_search_evaluation} shows ablation studies of the Product Search module, along with comparisons with a simple tf.idf baseline. In particular, after the 3rd turn of dialog, we selected top-5 products with highest probability from the list returned by Product Search module, and calculated recall, precision and F1 value against the ground-truth purchased product. We can see that the end-to-end search significantly improved the search recall by 4.69 times over the tf$\cdot$idf baseline. Improvement induced by combining text and attribute embedding suggests the benefits of combining product text and attributes in search task.

\subsection{Dialog Generation Method Evaluation}
We next conducted ablation studies on the data construction method. We evaluated the effectiveness of each component using the performance of State Tracker task. For each configuration in Table \ref{dataset_evluation}, we trained the State
Tracker module with corresponding dataset, and reported the performance on a manually prepared test set. As shown in the table, the module performance degrades without syntax analysis since redundant phrase (e.g. time, location) won't be removed from the utterance. Similarly, module performance degrades without paraphrase since language variance will be weakened. These suggest that both removing redundancy with syntax and increasing variance with paraphrase are effective to improve the training dataset quality.

\begin{table}[!tp]
    \centering
    \small
    \resizebox{\linewidth}{!}{
        \begin{tabular}{|l|ccc|}
        \hline 
        \bf{Dataset} & \bf{Attr} & \bf{Value} & \bf{Intent}  \\
        \hline 
         UT w/o Syntax \& Paraphrase  &  72.53  & 65.98  & 88.53   \\
         UT w/o Syntax                &  85.71  & 79.61  & 98.14   \\
         UT w/o Paraphrase            &  87.51  & 75.12  & 97.55   \\ 
        \hline 
         UT                   &  \bf{95.72}  & \bf{90.20}  & \bf{99.69}    \\
        \hline 
        \end{tabular}
    }
    \caption{The effectiveness of utterance generation methods for utterance understanding. The numbers in the table are F1 scores. We can see that both syntax and paraphrase improve the dialog data quality.}
    \label{dataset_evluation}
\end{table}

\begin{table}[!tp]
    \centering
    \small
        \begin{tabular}{|l|ccc|}
        \hline 
        \bf{Method} & \bf{Coherence}   & \bf{Fluency} & \bf{Approp.} \\
        \hline 
         TC-bot        &  2.98  & 3.42  & 3.20   \\
         ConvSearch    &  3.58  & 3.54  & 3.66   \\ 
        \hline 
        \end{tabular}
    \caption{Human Evaluation Result. Approp. is short for Appropriateness.}
    \label{human_evaluation}
     
\end{table}

\subsection{Human Evaluation}
We also performed human evaluations on system responses. For each method, we collected 100 dialogs and asked three workers to evaluate them with three metrics: coherence, fluency and appropriateness. All metrics have five grades: from 1(worst) to 5(best), where 3 denotes `good'. As shown in Table \ref{human_evaluation}, ConvSearch outperforms the baseline model in all three metrics.

\section{Conclusion and Future Work}
In this work, we built an end-to-end conversation search system for online shopping, where we deeply combined the dialog and search system with multi-task learning. In particular, our product search module leverages both product attribute and text to retrieve products, which mitigates the imperfect product schema/knowledge challenges.
To address issue of lacking in-domain dialog dataset, we proposed a dataset transfer method and constructed shopping dialog dataset from user search behavior data and existing dialogs of similar domain. The proposed dataset construction method lowers the cost, making it possible to scale-up to broader use scenarios.

We will leave it to future work to expand the methodology across more shopping categories, and broader use scenarios such as clinical conversations and customer service, etc.

\section*{Acknowledgments}
Hao He is supported by the National Key Research and Development Program of China under Grant 2018YFC0830400, the Basic Research Project of Shanghai Science and Technology Commission under ECNU-SJTU joint Grant 19JC1410102, the Shanghai Municipal Science and Technology Major Project under Grant 2021SHZDZX0102. This research is partially supported by the Shanghai Science and Technology Innovation Action Plan under Grant 20511102600.

\bibliography{emnlp2021,custom}
\bibliographystyle{acl_natbib}

\end{document}